\documentclass[letterpaper, 10 pt, conference]{ieeeconf}  

\IEEEoverridecommandlockouts                              

\usepackage{cite}
\usepackage{amsmath,amssymb,amsfonts}
\usepackage{algorithmic}
\usepackage{graphicx}
\usepackage{textcomp}
\usepackage{xcolor}
\usepackage{url}
\usepackage{siunitx}

\usepackage{tabularx,booktabs}
\newcolumntype{C}{>{\centering\arraybackslash}X} 
\usepackage{lipsum}

\usepackage{subcaption}

\title{\LARGE \bf
Deep Reinforcement Learning based Robot Navigation in Dynamic Environments using Occupancy Values of Motion Primitives
}

\author{Ne\c{s}et \"{U}nver Akmandor$^{1*}$, Hongyu Li$^{2\dagger}$, Gary Lvov$^{1\ddagger}$, Eric Dusel$^{1\mathsection}$ and Ta\c{s}k{\i}n Pad{\i}r$^{3}$
\thanks{*This research is supported by the National Science Foundation under Award
Number 1928654.}
\thanks{$^{1}$Department of Electrical and Computer Engineering, Northeastern University, Boston, MA, 02115, USA
        {\tt\small $^{*}$\{akmandor.n\},$^{\ddagger}$\{lvov.g\},$^{\mathsection}$\{dusel.e\} @northeastern.edu}}%
\thanks{$^{2}$Khoury College of Computer Sciences, Northeastern University, Boston, MA, 02115, USA
        {\tt\small $^{\dagger}$li.hongyu1@northeastern.edu}}%
\thanks{$^{3}$Institute for Experiential Robotics,
        Boston, MA, 02115, USA
        {\tt\small t.padir@northeastern.edu}}
}

\begin{document}

\maketitle
\thispagestyle{empty}
\pagestyle{empty}

\begin{abstract}

This paper presents a Deep Reinforcement Learning based navigation approach in which we define the occupancy observations as heuristic evaluations of motion primitives, rather than using raw sensor data. Our method enables fast mapping of the occupancy data, generated by multi-sensor fusion, into trajectory values in 3D workspace. The computationally efficient trajectory evaluation allows dense sampling of the action space. We utilize our occupancy observations in different data structures to analyze their effects on both training process and navigation performance. We train and test our methodology on two different robots within challenging physics-based simulation environments including static and dynamic obstacles. We benchmark our occupancy representations with other conventional data structures from  state-of-the-art methods. The trained navigation policies are also validated successfully with physical robots in dynamic environments. The results show that our method not only decreases the required training time but also improves the navigation performance as compared to other occupancy representations. The open-source implementation of our work and all related info are available at \url{https://github.com/RIVeR-Lab/tentabot}.

\end{abstract}

\section{INTRODUCTION}\label{sec:introduction}

Autonomous navigation in dynamic environments remains a challenging problem in robot motion planning \cite{cheng2018autonomous,cai2020mobile}. Although classical approaches such as graph-based \cite{dijkstra1959note,hart1968formal} and sampling-based \cite{lavalle1998rapidly,karaman2011sampling} methods can generate optimal paths, these techniques require prior map information while taking into account only static obstacles. This shortcoming limits their utility in real-world scenarios, such as autonomous driving within crowds \cite{bai2015intention,luo2018porca}. On the other hand, optimization-based methods \cite{liu2018mpc,lafmejani2021nonlinear} and reactive algorithms \cite{fox1997dynamic,quinlan1993elastic,ulrich2000vfh} are capable of avoiding dynamic obstacles. However, these approaches are limited as they require pre-determined motion models of the moving obstacles. ~\\

In recent years, Deep Reinforcement Learning (DRL) based approaches have shown remarkable results in mobile robot navigation \cite{tai2016robot,chen2017decentralized,pfeiffer2017perception,yan2020towards,doukhi2021deep,xu2021hierarchical} in unknown environments. Through observing their environment via sensors, trained agents in these works are able to explore unknown areas while avoiding obstacles. Although their efficiency depends on their training process, these data-driven approaches have two main advantages over classical methods. First, they gradually improve their performance during training by increasing the entropy of the defined state-action space. Second, they do not need an actual model of the environment to guide them to reach a successful policy. Despite their reported success, end-to-end learning methods require extensive training to achieve the desired navigation task \cite{guldenring2020learning,dugas2020navrep} due to large observation spaces. Without a model of their environment, these approaches try to estimate the necessary state values by interpreting high volumes of raw sensor data. To reduce the size of the input data, researchers mainly use Convolutional Neural Network (CNN) \cite{botteghi2021low} and Variational Auto-Encoder (VAE) \cite{dugas2020navrep} architectures. However, these methods add extra training process on top of the navigation tasks. ~\\

While the assumption of having the environment model is impractical for navigation tasks, calculating the kinematic model of a robot is straightforward. Using that, one can filter the workspace information based on the task. To the best of our knowledge, no previous work utilizes the robot's kinematic model in 3D workspace in order to reduce the size of the required observation space using the DRL architecture.

\subsection{Contribution}\label{sec:contribution}

To improve the efficiency of the DRL methods for robot navigation in dynamic environments, our contributions in this paper include:
\begin{itemize}
\item Synthesizing computationally efficient trajectory occupancy values, considering the robot's kinematics model in 3D workspace. We present their structural variants as inputs to different neural networks in the DRL architecture and analyze their effects on navigation performance.
\item Analyzing the performance of our occupancy representations by objectively comparing them with other state-of-the-art data structures within our presented framework. The policy of the robot is trained in challenging and various physics-based simulation environments.
\item Validating our presented methods and navigation framework via simulation and real-world tests on two different mobile robots.
\item Open-sourcing our work and results by providing our implementation and benchmark data in an code repository, which can be reached at \url{https://github.com/RIVeR-Lab/tentabot}. 
\end{itemize}

\section{RELATED WORK}\label{sec:related_work}

In the literature of robot navigation, several methodologies are verified to be useful in the presence of moving obstacles. Given the trajectory estimation of dynamical objects, classical methods, such as graph-based \cite{fraichard1994car} and optimization-based \cite{schoels2020ciao,lindqvist2020nonlinear}, are able to find robust and collision free trajectories. While their mathematical guarantees make them reliable over data-driven methods, classical methods suffer from computational complexity when the condition space, i.e. the number of moving obstacles, is scaled up. Another drawback of these classical approaches is the assumption of having future trajectory information of dynamic obstacles. Although there are data driven methods \cite{alahi2016social,shafiee2021introvert} that successfully predict human walking trajectories, they cannot be generalized for more complex motion models where the intentions of real-world agents can abruptly change. ~\\

Instead of assuming to know the agents' future trajectories, DRL based studies \cite{chen2017decentralized,chen2017socially,everett2018motion,chen2019crowd,sathyamoorthy2020densecavoid,tan2020deepmnavigate,kastner2020deep,everett2021collision} intrinsically estimate them by including the available obstacle data in their observation space. Considering the progress in pattern recognition and image processing, extracting the necessary information such as position, velocity, and each object's bounding box from the sensor data are more feasible than predicting obstacles' trajectories. Yet, the quality of the data extraction for each obstacle determines the success of the learned policy. Having a limited processing time, these works are similarly restrained by the number of obstacles. Although the aforementioned papers present dynamic obstacle avoidance in crowds (up-to 90 agents as in \cite{tan2020deepmnavigate}), this can only be achieved by using the available obstacle trajectories within their simulation environment. In contrast, \cite{chen2017socially,kastner2020deep,everett2021collision} estimate pedestrian velocities from the 2D range scan data and RGB images of their physical robot. Despite their results, there remains a huge feasibility gap between simulation and actual experiments. ~\\

To avoid dynamic obstacles without estimating trajectories or extracting current states, robots can be trained by supervised policies with imitation learning \cite{hussein2017imitation}. Ross et al. present in \cite{ross2013learning} a learning based Micro Aerial Vehicles (MAVs) controller which is trained with human expert data. At each time step, camera data are mapped to a set of features as the observation. Then, the reactive control policy is learned by the neural network to resemble the expert's performance. To bypass the tedious supervised data collection process, Pfeiffer et al. use a motion planner as their expert operator to train their network model in \cite{pfeiffer2017perception}. Their observation space includes raw data from the LiDAR sensor and 3D target information. One particular disadvantage of such methods is that the learned policy is limited by the supervised system's capabilities. ~\\

Without the aforementioned inputs (predicted obstacle trajectory, observable obstacle states, and expert policy), another group of work \cite{tai2016robot,tai2017virtual,xie2018learning,long2018towards,zeng2019navigation,guldenring2020learning,gao2020deep,liu20203d,dugas2020navrep,perez2020robot,patel2021dwa,song2021multimodal} employs only occupancy observations to determine optimal navigation policy with DRL. To manage the elevated complexity, the researchers focus on finding efficient representations and neural network (NN) modules to encapsulate the necessary information to achieve the navigation task. Table \ref{tab:literature} summarizes most relevant approaches to our work by listing their observation data, NN architectures, and reinforcement learning methods.
\begin{table}[ht]
    \caption{List of observation data, NN architectures and RL methods in DRL based robot navigation literature.}
    \label{tab:literature}
    \centering
    \resizebox{\columnwidth}{!}{%
    \begin{tabular}{|l|c|c|c|}
    \hline
    Ref. & Data (*Processed, $^{\dagger}$Stacked) & NN Architecture & RL Method\\
    \hline
    \cite{tai2016robot} & depth image & 2D\_CNN+FC & DQN \\
    \hline
    \cite{tai2017virtual} & laser scan, target, action & FC & DDPG, \\
    \hline
    \cite{xie2018learning} & laser scan$^{\dagger}$, target, action & 1D\_CNN+FC & DQN \\
    \hline
    \cite{long2018towards} & laser scan$^{\dagger}$, target, action & 1D\_CNN+FC & PPO \\
    \hline
    \cite{zeng2019navigation} & laser scan*, action & FC & PPO \\
    \hline
    \cite{guldenring2020learning} & laser scan*$^{\dagger}$, waypoints & 1D\_CNN+FC, & PPO \\
    & & 2D\_CNN+FC & \\
    \hline
    \cite{gao2020deep} & laser scan, target, action & FC & TD3 \\
    \hline
    \cite{liu20203d} & laser scan, depth image, & \{1D+2D\}\_CNN+FC & SAC \\
    & waypoints, action & & \\
    \hline
    \cite{dugas2020navrep} & laser scan, target, action & VAE+LSTM+1D\_CNN+FC, & PPO \\
    & & VAE+LSTM+2D\_CNN+FC & \\
    \hline
    \cite{perez2020robot} & laser scan, waypoints, target & 1D\_CNN+FC & SAC \\
    \hline
    \cite{patel2021dwa} & laser scan*$^{\dagger}$, target*$^{\dagger}$, action$^{\dagger}$ & 2D\_CNN+FC & PPO \\
    \hline
    \cite{song2021multimodal} & laser scan, depth image, action$^{\dagger}$ & \{1D+2D\}\_CNN+FC & D3QN \\
    \hline
    \hline
    Ours & laser scan*$^{\dagger}$, depth image*$^{\dagger}$, & FC,1D\_CNN+FC & PPO \\
    & target, action & 2D\_CNN+FC & \\
    \hline
    \end{tabular}
    }
\end{table}

\section{APPROACH}

\subsection{Problem Definition and Navigation Framework}

We define our problem as a Partially Observable Markov Decision Process (POMDP) \cite{kaelbling1998planning} since neither geometric models of static obstacles nor the trajectory of mobile agents are known prior to the navigation. The POMDP model $M=(S,A,T,R,\Omega,O,\gamma)$ consists of state space $S$, action space $A$, state transition function $T$, reward function $R$, finite set of observations $\Omega$, observation function $O$ and discount factor $\gamma \in [0,1)$. Given the current state $s \in S$ and the selected action $a \in A$, the state transition function $T(s'|s,a)$ calculates the probability of the next state $s' \in S$. As the actual state of the robot is assumed to be unavailable in POMDP formulation, it is estimated based on the observation function $O(o|s',a)$ where $o \in \Omega$. At each time step, the agent chooses an action based on the recent observations and receives an immediate reward $r=R(s,a)$ from the environment. The main aim of this formulation is to find the policy function $\Pi(a|s)$ that maximizes the cumulative discounted reward. Therefore, the reward function needs to be designed to lead the agent toward the goal state. ~\\

\begin{figure}[ht]
\centering
\includegraphics[width=3in]{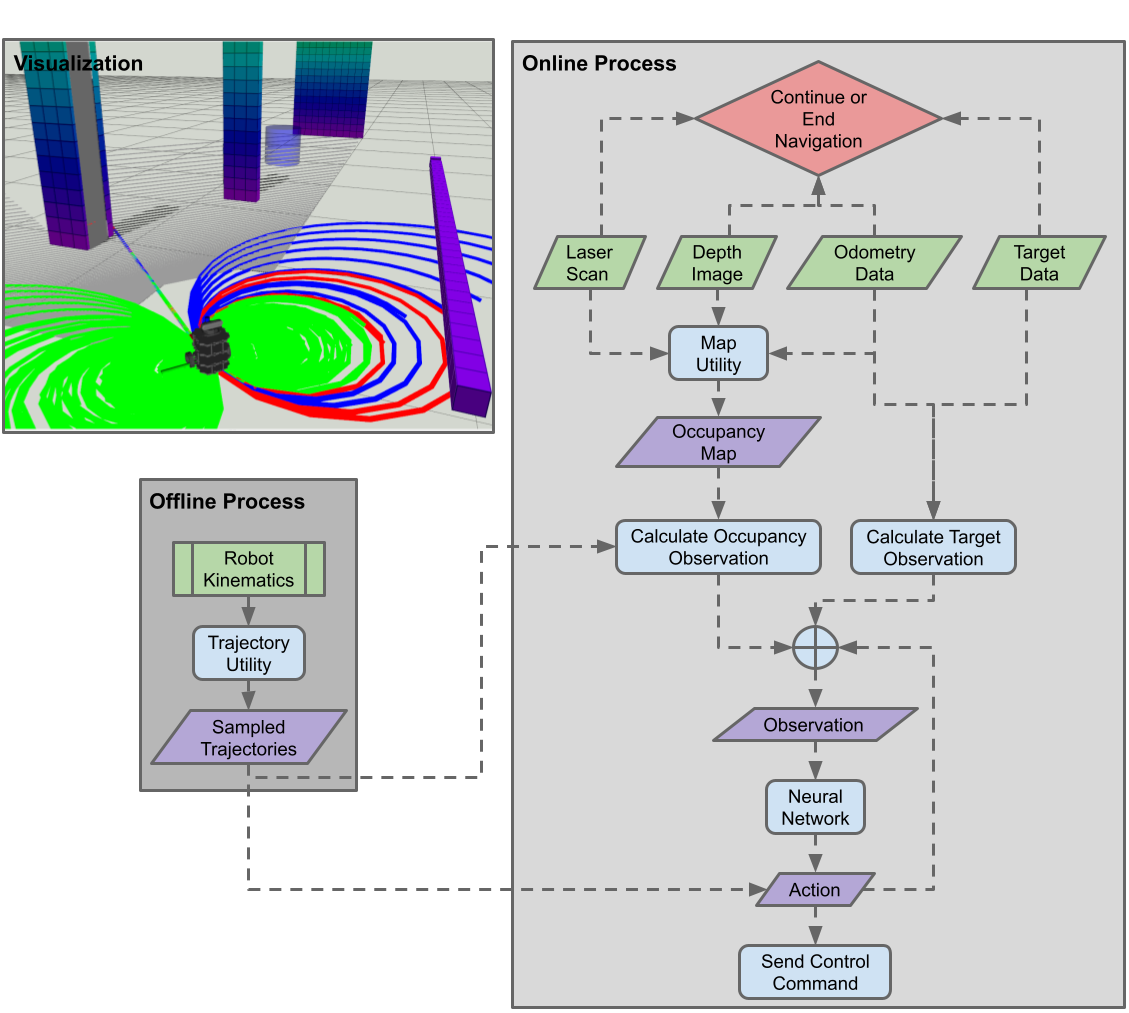}
\caption{Our navigation framework consists of two main stages: (1) In the Offline stage, trajectories are sampled using the robot's kinematic model. A 3D grid structure, composed by voxels, is formed around the robot. The voxels are classified as either "Priority" or "Support" based on their proximity to the nearest sampling point. (2) During the Online stage, sensor data are mapped into classified voxels to enable the calculation of occupancy values for each trajectory. Using these observations as the input, the neural network outputs a discrete action. The online stage continues until a terminal state (goal or collision) is reached.}
\label{fig:framework}
\end{figure}

Our proposed framework is given in Fig. \ref{fig:framework}. Our task, robot and environment-specific observation space, action space and reward function are described in sections \ref{sec:observation}, \ref{sec:action} and \ref{sec:reward} respectively. The state estimation and the policy function are jointly estimated by the neural network whose architecture is defined in Section \ref{sec:neural_network}.

\subsection{Action Space}\label{sec:action}

Having linear $v \in [0, v^{max}]$ and angular $\omega \in [\omega^{min}, \omega^{max}]$ velocity limits of the robot's controller, we uniformly sample both spaces by $n^v$ and $n^{\omega}$ respectively. Providing initial states and discretized values of the action space as an input to the kinematic model, we form $n^k=n^vn^{\omega}$ trajectories. Hence, each trajectory $j$ has one-to-one mapping to an action tuple $(v_j, \omega_j)$. At each time step, the action index from the deep neural network output determines the velocity command. ~\\

Throughout this paper, we use 2D vector space to represent our actions $a=[v,\omega] \in \mathbb{R}^2$ since we perform our simulations on a differential drive robot controlled by linear $v$ and angular $\omega$ velocity commands. However, our approach is readily applicable to 3D navigation and position-controlled systems, such as uninhabited aerial vehicles (UAVs) as well. In that case, the action space is defined as a 3D vector on the selected trajectory as in \cite{akmandor20203d,akmandor2021reactive}.

\subsection{Observation Space}\label{sec:observation}

Instead of including raw sensor information, such as laser scan as in \cite{tai2017virtual,pfeiffer2017perception,xie2018learning,long2018towards,gao2020deep,liu20203d,dugas2020navrep,perez2020robot,kastner2020deep,song2021multimodal} or depth image as in  \cite{tai2016robot,tai2018socially,sathyamoorthy2020densecavoid,liu20203d,song2021multimodal}, our observations utilize the occupancy information of the pre-sampled trajectories. To efficiently evaluate these trajectory values at each time step, the framework in \cite{akmandor20203d,akmandor2021reactive} is adopted in our DRL based architecture. The 3D grid structure, formed by voxels, is centered at the robot's frame $R$ and initialized prior to navigation. We sample each trajectory $j$ by $n^{T_j}$ points $^{R}p_i=(x,y,z)$ and keep them in a set $T_j = \{^{R}p_i \quad | \quad i = 1,...,{n^{T_j}}\}$. Each voxel is defined as $\psi_k=(u_k, \beta_k, m_k, c_k)$, where $u_k$ is the index that points out the $k^{th}$ voxel position in the linearized array $A_p$ which keeps coordinate positions of voxel centers. $\beta_k$ is the user-defined weight for the corresponding voxel. $m_k$ is the index of the closest sampling point on the trajectory to the $k^{th}$ voxel. $c_k$ indicates whether it is a Priority ($c = 1$) or Support ($c = 0$) voxel. The set of ``Support'' $S^{\psi}_j = \{\psi_{k} | k = 1,...,{n^{S^{\psi}_j}}, {\tau^{P^{\psi}}} < || A_p(u_k) - T_{m_k} || < {\tau^{S^{\psi}}}\}$ and ``Priority'' $P^{\psi}_j = \{\psi_{k} | k = 1,...,{n^{P^{\psi}_j}}, || A_p(u_k) - T_{m_k} || < {\tau^{P^{\psi}}}\}$ voxels inside the robot-centered 3D grid are extracted based on the user-defined thresholds ${\tau^{S^{\psi}}}$ and ${\tau^{P^{\psi}}}$ respectively. ~\\

Inspired by their formulation of two heuristic functions, which are ``Clearance'' and ``Nearby Clutter'', we define our distinct occupancy function $H^{occ}$  to represent both information as a single value. We define $H^{occ}$, as shown in Eq. \ref{eqn:occupancy}, using similar notations in \cite{akmandor2021reactive}. For each sampling point $i$ in trajectory $j$, sum of voxel weights $W_{j}$ is calculated by Eq. \ref{eqn:beta_sum}. At each time step, linearized array $A_{\sigma}$ which consists of occupancy values inside the 3D grid are updated by the recent sensor data. The crash index $u^{crash}$ is determined by the first occupied Priority voxel along the trajectory. Having $A_{\sigma}$ and $u^{crash}$ for each trajectory $j$, the weighted sum of voxel weights $W_{j}^{scaled}$ is calculated by Eq. \ref{eqn:occupied_beta_sum}. Then, the occupancy value $H_j^{occ}$ is obtained by $W_{j}^{scaled}$ divided by the multiplication of $W_{j}$ with the maximum occupancy value of a voxel $\sigma^{max}$. By definition, $H_j^{occ} \in [0,1]$.
\begin{align}\label{eqn:occupancy}
    H_j^{occ} = \frac{W_j^{scaled}}{\sigma^{max} W_j}
\end{align}
\begin{align}\label{eqn:beta_sum}
    W_{j} = \sum_{k=1}^{{n^{S^{v}_j \cup P^{v}_j}}} \beta_{k},
\end{align}
\begin{align}\label{eqn:occupied_beta_sum}
    W_{j}^{scaled} = \sum_{i=1}^{{n^{S^{v}_j \cup P^{v}_j}}} &\alpha_{m_k} \beta_k,
\end{align}
\begin{align}
    \nonumber
    \text{where} \quad &\psi_{k} \in S_j^{v} \cup P_j^{v}, \\
    \text{and} \quad &\alpha_{m_k} = 
    \begin{cases}
        \nonumber
        A_{\sigma}(m_k) & if \quad m_k < u^{crash}\\
        \sigma^{max} & else.
    \end{cases}
\end{align}

Similar to \cite{tai2017virtual,xie2018learning,long2018towards,gao2020deep,dugas2020navrep}, we include the target data $o^{target}=[d^{target},\theta^{target}] \in \mathbb{R}^2$ and previous action $o^{action}=[v^{pre},\omega^{pre}] \in \mathbb{R}^2$ in our observation space. The first field of the target data $d^{target} \in \mathbb{R}^+$ is defined as the Euclidean distance between the robot and the target. The second field is defined as the yaw angle $\theta^{target} \in [-\pi,\pi]$ to the target with respect to robot's frame $R$. ~\\

\begin{figure}[ht]
\centering
\includegraphics[width=3in]{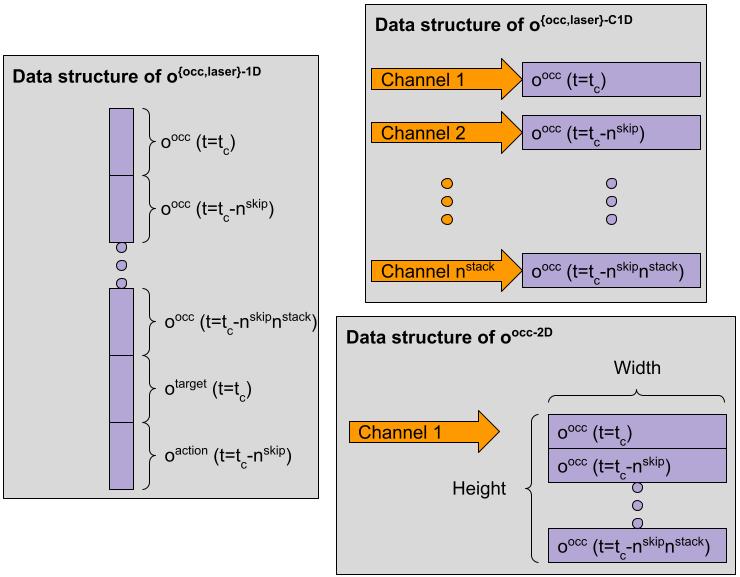}
\caption{Different structures of the observation data are generated by either laser range or trajectory occupancy values. Left: Data from different time steps are concatenated into a 1D array. Right-Top: For each time step, 1D array of data are passed into a channel. Right-Bottom: Each row of the 2D array structure is formed by the data at its corresponding time step. The height shows the number of stacked data from previous time steps.}
\label{fig:obs_struct}
\end{figure}

To learn the dynamic characteristics of obstacles, analogous to \cite{xie2018learning,long2018towards,guldenring2020learning,patel2021dwa}, we stack occupancy data from $n^{stack}$ previous time steps as shown in Fig. \ref{fig:obs_struct}. Since merely milliseconds pass between successive steps, we observe better results when we skip $n^{skip}$ time steps when stacking the data to add into the observation space. The structure of the stacking depends on the NN architecture that we input our observations. ~\\

To utilize the fully connected (FC) layers as on the left in Fig. \ref{fig:nn_arch}, we obtain our occupancy observations by concatenating one after another as in $o^{occ-1D} = [H_{1t}^{occ},...,H_{n^kt},H_{1(t-n^{skip})}^{occ},...,H_{n^k(t-n^{skip}n^{stack})}^{occ}] \in \mathbb{R}^{n^kn^{stack}}$. We consider the occupancy observation at each time step $o^{occ-C1D} = [H_{1}^{occ},...,H_{n^k}] \in \mathbb{R}^{n^k}$ for the input to each channel in 1D-CNN architecture. Therefore for $n^{stack}$ time steps, we have $n^{stack}$ channels of data where each row belongs to a particular time step. For the input of 2D-CNN network, we present our occupancy data in a 2D array structure. Different from the 1D-CNN case, we form each column to represent the data at each time step. Therefore, the time stack of data concatenated in the row direction and the observation space becomes $o^{occ-2D} \in \mathbb{R}^{n^k \times n^{stack}}$.

\subsection{Reward Function}\label{sec:reward}

We design our reward function as in Eq. \ref{eqn:reward} to supervise the robot to reach the goal and avoid collision with obstacles. We set a positive reward $\mu^{goal}$ when the robot reaches the goal. If the episode ends by either passing a threshold $\tau^{fail}$ value of a distance to the closest obstacle $d_t^{obs}$ or reaching the maximum number of steps $n^{max\_ep\_ts}$ per episode, we penalize it by a negative reward $\mu^{fail}$. To reduce the training time that the robot might inefficiently spend to explore its state space, we define a step reward function $\mu_t^{step}$ to evaluate the states other than the end cases.
\begin{align}\label{eqn:reward}
    Reward &= 
    \begin{cases} 
        \mu^{goal} \textbf{  if  } d_t^{target} < \tau^{target}\\
        \mu^{fail} \textbf{  else if  } n_t > n^{max\_ep\_ts} \textbf{ or } d_t^{obs} < \tau^{fail}\\
        \mu_t^{step} = \mu_t^{target} + \mu_t^{step\_pen} \textbf{  else} \\
    \end{cases} \\
    \label{eqn:reward2}
    &\textbf{where  } \mu_t^{target} = \alpha^{target} (d_{t-\Delta t}^{target} - d_t^{target}) \\
    \label{eqn:reward3}
    &\quad\quad\quad \mu_t^{step\_pen} = \frac{\alpha^{step\_pen}}{n^{max\_ep\_ts}}  
\end{align}
The first part of the step function $\mu_t^{target}$ is calculated by taking the difference between current and previous time steps' target distance and multiplying it with the user-defined scalar $\alpha^{target}$ to adjust the reward/penalty weight on the navigation. The target distance is calculated with respect to the robot's frame $R$. Based on the sign of the distance difference, the function either penalizes or rewards the current state. The second part of the step function $\mu_t^{step\_pen}$ penalizes the robot for not reaching the goal. The penalty value $\mu_t^{step\_pen}$ is fixed for each time and determined by the user-defined scalar $\alpha^{step\_{pen}}$ and $n^{max\_ep\_ts}$ as in Eq. \ref{eqn:reward3}. Although the robot may tend to get stuck at local minima because of the greedy evaluations of our reward design, including previous action in observations helps robot to escape from dead spots by forcing it to choose different actions.

\subsection{Neural Network Architectures}\label{sec:neural_network}

\begin{figure}[ht]
\centering
\includegraphics[width=3in]{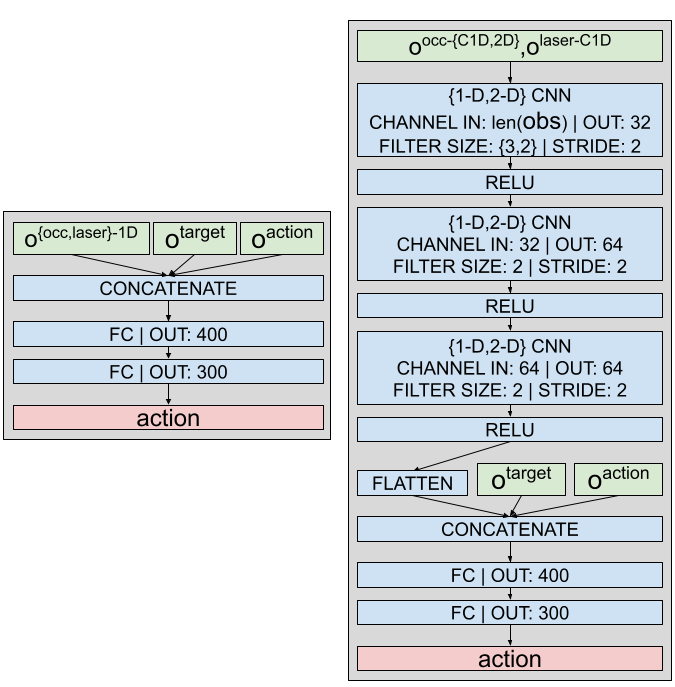}
\caption{Neural network architectures where the observation inputs are shown in green, network layers in blue and action output in red. Left: 2 FC layers.The observations $o^{occ-1D},o^{laser-1D}$ are concatenated with target $o^{target}$ and action $o^{action}$ observations before going into FC layers. Right: 3-layer CNN followed by 2 FC layers. The occupancy observations $o^{occ},o^{laser}$ are fed into either 1D or 2D CNN depending on their data structures. The target $o^{target}$ and action $o^{action}$ observations are concatenated with the output of the CNN and then input into the FC layers.}
\label{fig:nn_arch}
\end{figure}

To analyze the effect of the input structure of our trajectory value observations on navigation performance, we train the robot using three different neural network architectures as shown in Fig. \ref{fig:nn_arch}.
The left side of Fig. \ref{fig:nn_arch} displays the fully connected (FC) layers which take the occupancy, either $o^{occ-1D}$ or $o^{laser-1D}$, target $o^{target}$ and action $o^{action}$ observation inputs. We concatenate these three observations into a 1D array and feed it into the first layer of the FC network. The output of the network is mapped into one of the discrete actions defined in \ref{sec:action}. We also implement two different CNN networks both of which consist of 3 CNN layers followed by 2 FC layers as shown in the right side of Fig. \ref{fig:nn_arch}. The only difference between both networks is the input structure of the observations as defined in Section \ref{sec:observation} and visualized in Fig. \ref{fig:obs_struct}.

\subsection{Implementation Details}\label{sec:implementation}

The core implementation of our work is built on the framework presented in \cite{akmandor20203d,akmandor2021reactive}, which is developed in ROS architecture. We utilize that coding base to construct the necessary structures, such as 3D grid, voxels and heuristic functions, generate pre-sampled trajectories, and map the occupancy data coming from the sensors (RGB-D camera and 2D LiDAR). The Gazebo simulator is used to form physical environments, simulate and control realistic robots, such as Turtlebot3 and Hello Robot's Stretch. We create the dynamic obstacles as groups of walking pedestrians using the implementation of the work in \cite{okal2014towards}. We also use Gym \cite{openai}, which is an open-source Python library for developing and benchmarking reinforcement learning algorithms. To construct a Gym environment within the ROS architecture, we utilize the ROS package in \cite{openai_ros}. ~\\
\\
We use Proximal Policy Optimization (PPO) \cite{schulman2017proximal} to train our robot for the navigation task within dynamic environments. PPO is a policy gradient method which is suitable for both continuous and discrete action spaces. Notably, it provides stability by determining the maximum step size for the exploration, so that the deviation between the policy iterations is bounded by a trust region. We employ the Stable-Baselines3 \cite{stable-baselines3} implementation of PPO, where they provide useful tools for customizing neural network architecture and monitoring the training process.

\section{RESULTS}\label{sec:results}

\subsection{Training}\label{sec:training}

\begin{figure}[ht]
\centering
\includegraphics[width=3in]{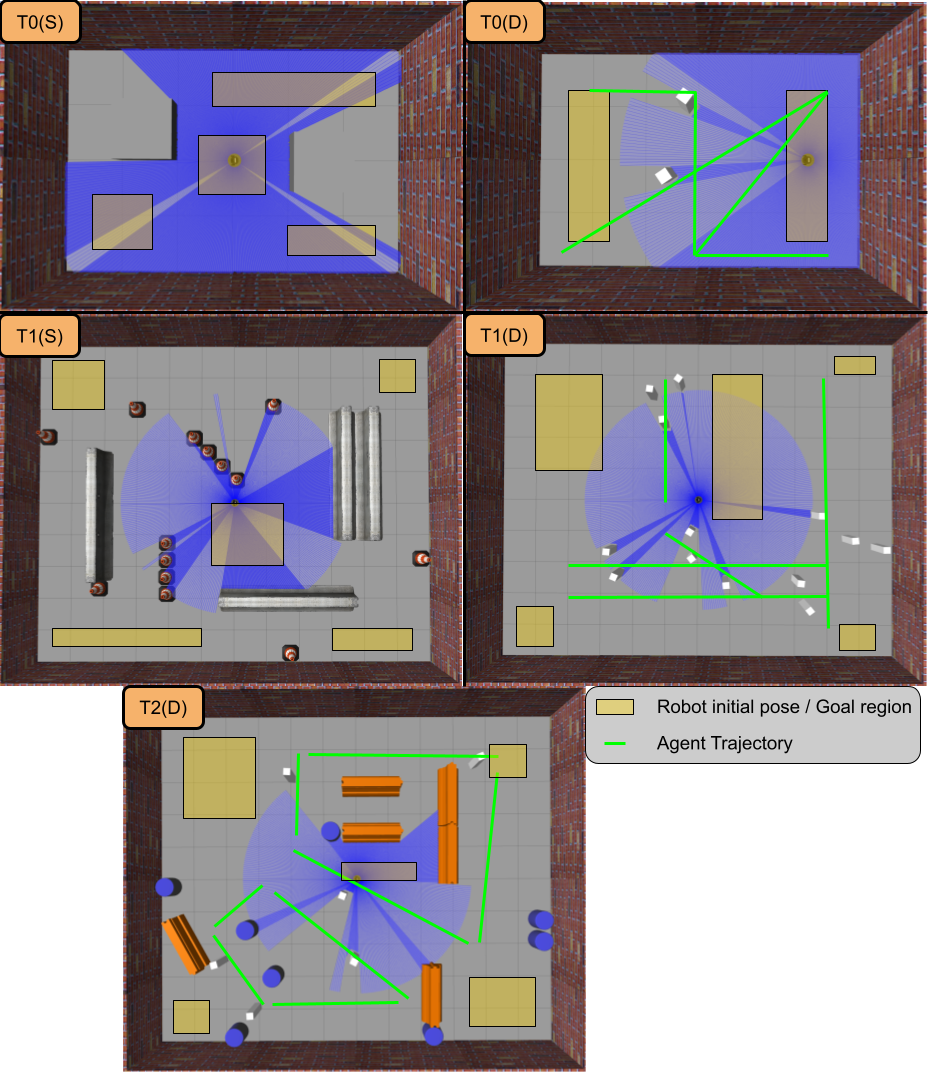}
\caption{Training environments: $T0(S)$ and $T1(S)$ contain only static objects. $T0(D)$ and $T1(D)$ labeled maps include only dynamic objects in the shape of rectangular cuboids. $T2(D)$ includes both static and dynamic obstacles. Robot and goal positions are randomized inside of the yellow areas.}
\label{fig:training_envs}
\end{figure}

To train our robot in simulations, we create five different environments in Gazebo as shown in Fig. \ref{fig:training_envs}. The environments labeled as $T0(S)$ and $T1(S)$ contain only static obstacles such that $T0(S)$ has two cubic (\SI{1}{m} diameter) objects and $T1(S)$ has more distinct objects. On the other hand, $T0(D)$ and $T1(D)$ labeled maps include only dynamic objects in the shape of rectangular cuboids (width: \SI{0.2}{m}, length: \SI{0.2}{m}, height:\SI{1}{m}). Similar to static maps, the number of obstacles in $T1(D)$ is higher than $T0(D)$ with 13 dynamic obstacles as opposed to 2. $T2(D)$ includes both static and dynamic obstacles. The area of $T0(S)$ and $T0(D)$ are \si{6x4}\;\si{m^2} while $T0(S)$, $T1(S)$ and $T2(D)$ are \si{12x10}\;\si{m^2}.

At the beginning of each training episode, the robot is initialized at a random position inside the yellow areas of the respective map in Fig. \ref{fig:training_envs}. The goal position is also randomly selected from these areas, without coinciding with the initial area of the robot. Similar to \cite{gao2020deep}, we use hierarchical training methodology starting from $T0(S)$, which has a small area and only contains static obstacles. Whenever we observe a convergence in cumulative rewards, we switch training in a more complex map in the order of $T1(S)$, $T0(D)$, $T1(D)$ and $T2(D)$. During our experiments, we found that the hierarchical process can remarkably improve the convergence rate compared to training only within a single complex map. ~\\

Our training benchmark includes seven different approaches which differ from each other by the occupancy data structures (raw laser scan, image representation \cite{guldenring2020learning}, polar representation\cite{dugas2020navrep}, our occupancy representation which is labeled as Tentabot) and neural network architectures (FC, 1DCNN\_FC, 2DCNN\_FC). To speed up the computation without effecting the performance for laser\_FC and laser\_1DCNN\_FC, we reduce the total number of laser scans from $360$ to $90$ by filtering them equidistantly. For the fairness of comparisons as having similar number of occupancy values in our variants, we linearly sample lateral and angular velocities by $5$ and $21$ respectively. That leads to $105$ trajectories with their occupancy values. Since our occupancy values are defined in the range of $[0,1]$, we also normalize the laser scan data. To benchmark our proposed value function, we use two different representations of the laser data: the 2D image representation from \cite{guldenring2020learning} labeled as laser\_image and the polar image representation from \cite{dugas2020navrep} labeled as laser\_rings. ~\\

\begin{figure}[ht!]
\centering
\includegraphics[width=3in]{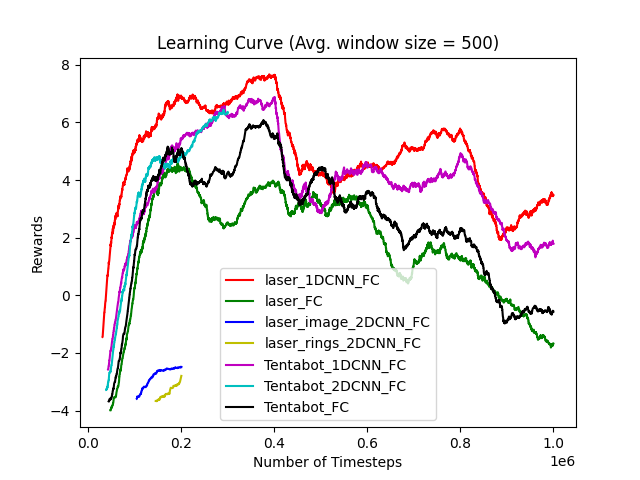}
\caption{Cumulative learning curve of the training data.}
\label{fig:training_result}
\end{figure}

We train all approaches using the aforementioned hierarchical methodology, where each training environment is run for 200k time steps. Since the cumulative rewards of laser\_image and laser\_rings could not reach positive values, even after 200k training steps in the first map $T0(S)$, we do not proceed with their training. Although the training performance of Tentabot\_2DCNN\_FC is as successful as Tentabot\_1DCNN\_FC, we also do not continue its training after it clearly outperforms its image counterparts, to reserve our limited computational resources for other comparisons. The overall training results in Fig. \ref{fig:training_result} show that higher cumulative rewards are achieved when the observations include our proposed occupancy value $o^{occ}$ over $o^{laser}$. Although the laser\_1DCNN\_FC converges faster than Tentabot\_1DCNN\_FC, they both gain competitive rewards during the training process. Similarly, there is no clear winner between laser\_FC and Tentabot\_FC. Here, we should point out that our representation has an important disadvantage against its laser counterparts. Although we use same network structures for each comparison, our occupancy representation has $75$ more values, as we stack $5$ previous observations, which causes less learning within the same amount of time. As a future work, we will address this issue by optimizing hyper-parameters, such as number of layers and number of neurons in these layers, of the neural network for each of these data structures.

\subsection{Testing}\label{sec:testing}

\begin{figure}[ht]
\centering
\includegraphics[width=\linewidth]{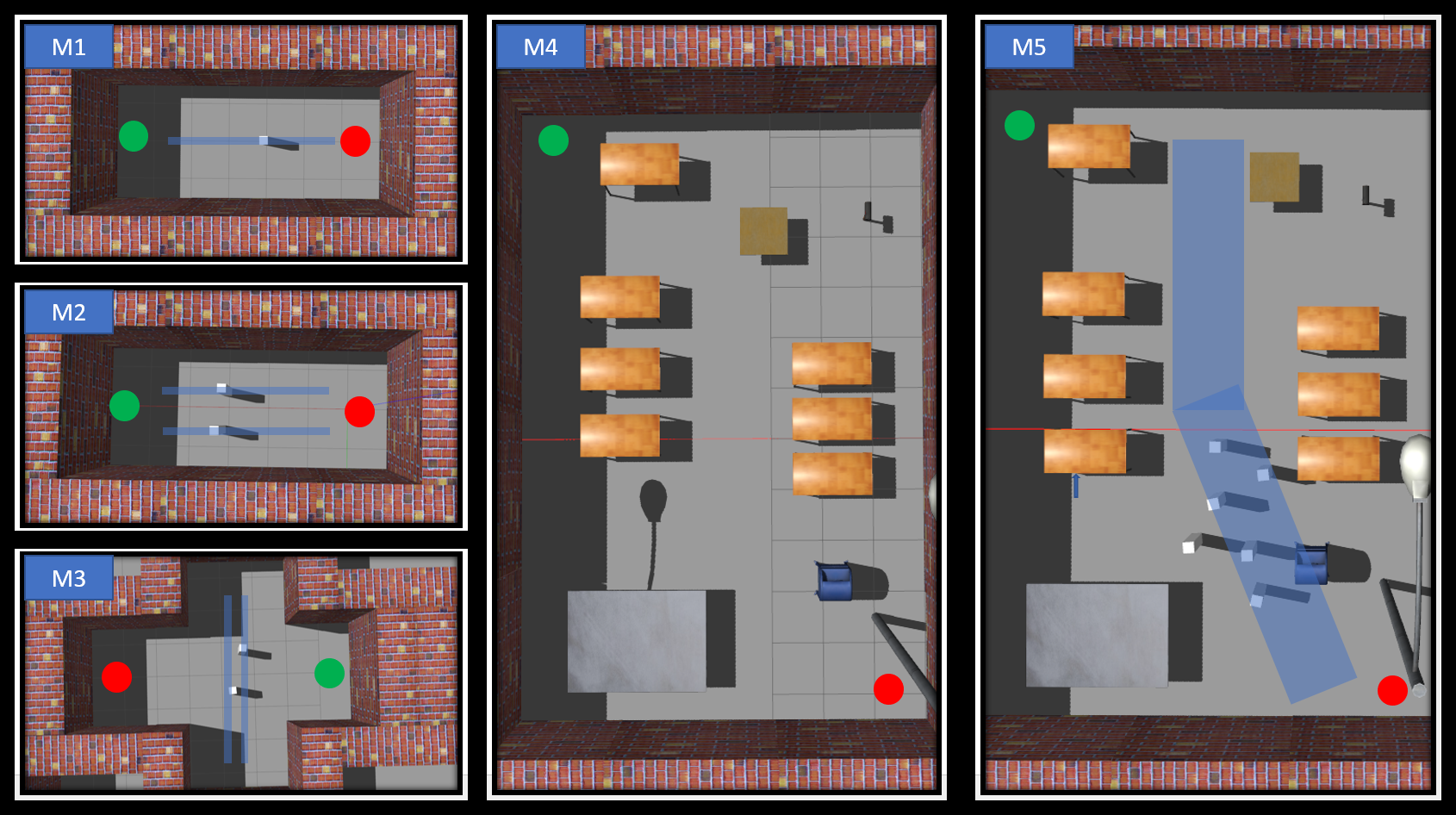}
\caption{Testing environments. The blue areas represent agent trajectories. The green dots represent initial point, and the red dots represent the goals.}
\label{fig:testing_envs}
\end{figure}

\begin{figure}[ht]
\centering
\vspace*{0.2cm}
\begin{subfigure}{.49\linewidth}
    \centering
    \includegraphics[width=\linewidth]{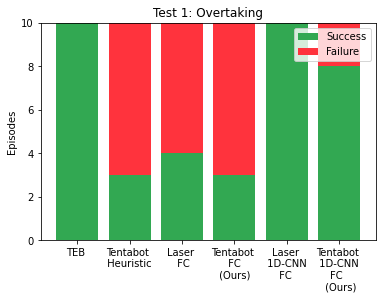}
    \caption{M1}\label{fig:image1}
    \label{fig:M1}
\end{subfigure}
\begin{subfigure}{.49\linewidth}
    \centering
    \includegraphics[width=\linewidth]{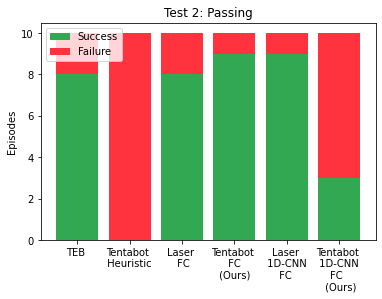}
    \caption{M2}
    \label{fig:M2}
\end{subfigure}
\begin{subfigure}{.49\linewidth}
    \centering
    \includegraphics[width=\linewidth]{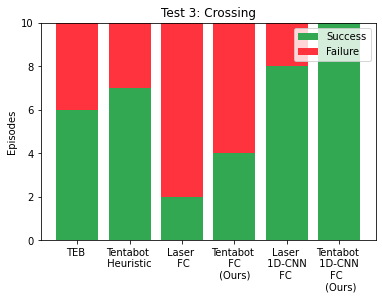}
    \caption{M3}
    \label{fig:M3}
\end{subfigure}
\begin{subfigure}{.49\linewidth}
    \centering
    \includegraphics[width=\linewidth]{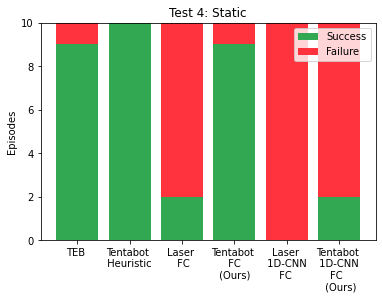}
    \caption{M4}
    \label{fig:M4}
\end{subfigure}
\begin{subfigure}{.49\linewidth}
    \centering
    \includegraphics[width=\linewidth]{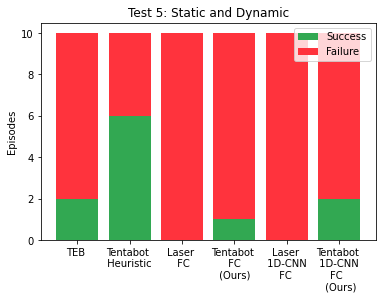}
    \caption{M5}
    \label{fig:M5}
\end{subfigure}
\caption{Testing results of different navigation approaches in 5 distinct environments. Approaches from left to right: TEB, Tentabot\_Heuristic, laser\_FC, Tentabot\_FC, laser\_1DCNN\_FC, Tentabot\_1DCNN\_FC. Each of the DRL based approaches is trained one million time steps which takes around 50 hours.}
\label{fig:testing_result}
\end{figure}

To compare the navigation policies trained by the different approaches, we perform simulation tests in five environments as shown in Fig. \ref{fig:testing_envs}. We create these environments with distinct static obstacles and map features different from the the training maps to demonstrate that our models are not overfitting. The testing environments include overtaking an agent moving towards the goal (Fig \ref{fig:M1}), overtaking agents moving towards the robot (Fig \ref{fig:M2}), avoiding agents which cross in front of the robot (Fig \ref{fig:M3}), crossing a complex static environment (Fig \ref{fig:M4}), and crossing complex static environment with random dynamic agents (Fig \ref{fig:M5}).

Based on the training results in Fig. \ref{fig:training_result}, we perform navigation tests using four successfully trained policies; laser\_FC, laser\_1DCNN\_FC, Tentabot\_FC, Tentabot\_1DCNN\_FC, and two reactive planners; Tentabot\_Heuristic \cite{akmandor20203d} and Timed Elastic Band (TEB) \cite{rosmann2013efficient}. We compare their performance in terms of navigation success rate (Fig \ref{fig:testing_result}). To test the robustness of each policy, we perform the same task ten times and calculate their mean values. The testing results show that the policies trained with our approach (Tentabot\_FC, Tentabot\_1DCNN\_FC, and Tentabot\_2DCNN\_FC) navigate with higher success rate and reach the goal faster than the laser\_FC, which utilizes normalized laser data. Moreover, Tentabot\_1DCNN\_FC outperforms all other approaches in the crossing environment (Fig \ref{fig:M3}), and matches with TEB in the complex static and dynamic environment (Fig \ref{fig:M5}). Even though laser\_1DCNN\_FC performs well in the training and the first three testing environments (Fig \ref{fig:M1}, Fig \ref{fig:M2}, Fig \ref{fig:M3}) where obstacles are structured, it performs poorly when the map consists of novel obstacles as in (Fig \ref{fig:M4}, \ref{fig:M5}). Our approach demonstrates greater performance in novel environments compared to the laser-based counterparts due to its embedded occupancy representation of the 3D workspace.
\subsection{Real-robot Testing}\label{sec:realrobot}
\begin{figure}[ht]
\vspace*{0.2cm}
\begin{subfigure}{.5\linewidth}
    \centering
    \includegraphics[width=.8\linewidth, height=1.8in]{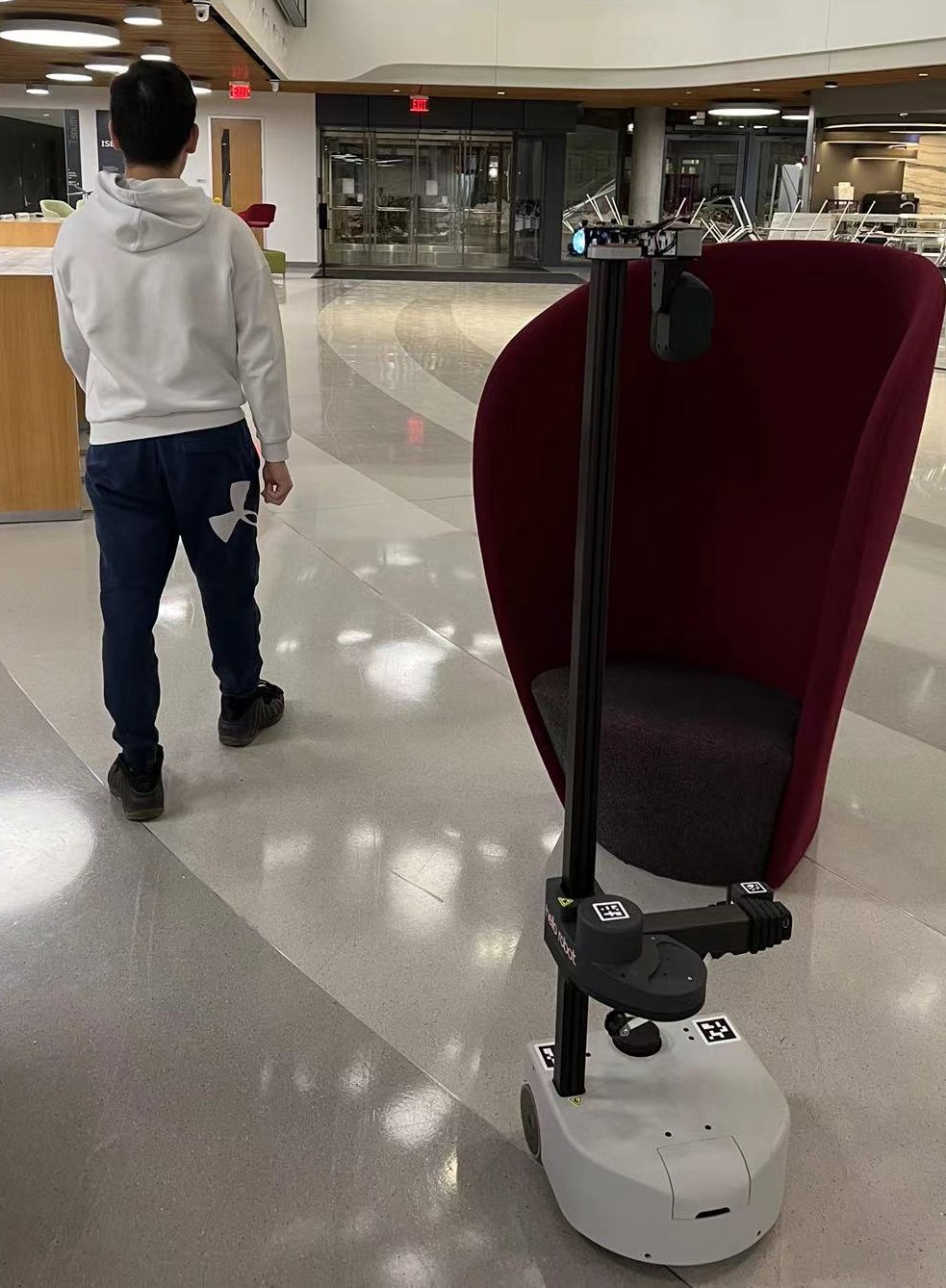}
    \caption{Real world scene}
    \label{fig:real_world_testing}
\end{subfigure}%
\begin{subfigure}{.5\linewidth}
    \centering
    \includegraphics[width=.8\linewidth, height=1.8in]{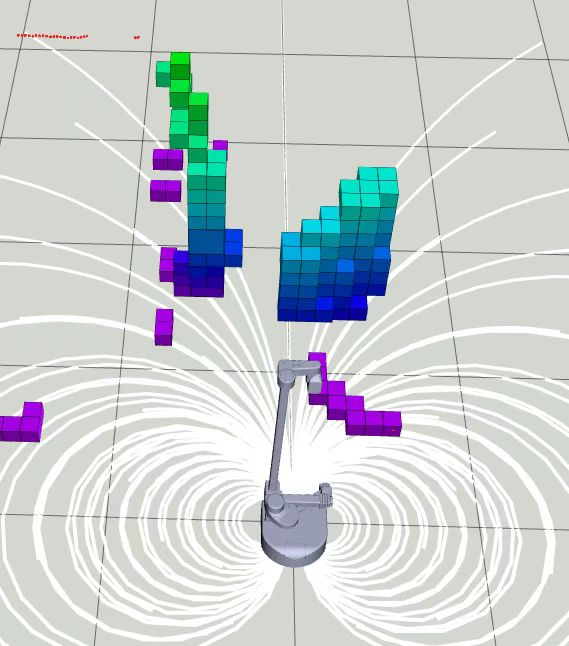}
    \caption{Visualization in RViz}
    \label{fig:real_world_testing_rviz}
\end{subfigure}
\caption{(a) We test our navigation approach in real world scenarios where static obstacles and dynamic agents are involved. (b) Our software runs in the robot's onboard computer where the real-time data, such as pre-sampled trajectories and occupied voxels, can be visualized.}
\end{figure}

We use the trained Tentabot\_FC policy on the Stretch from Hello Robot (Fig. \ref{fig:real_world_testing}) which is equipped with a Slamtex RPLiDAR A1, Intel RealSense D435i RGB-D Camera, and an onboard computer (Intel i5-8259U with 16GB RAM). We publish various goals in multiple environment settings (no obstacles, static obstacles, dynamic obstacles, and crowd) to test the performance of our work.
In real world experiments, our robot reaches the goal every time, even it has some physical contacts with dynamic agents but without colliding any static obstacle. Although we sometime observe that the robot circles around the goal, we infer that is due to the lack of parameter tuning in our reward function.

\section{CONCLUSIONS}\label{sec:conclusion}

In this paper, we introduce a DRL based robot navigation approach which uses occupancy values of pre-sampled trajectories as a part of the observation space. We present variants of this representation by utilizing different neural network architectures. To verify our approach, we implement our open-source work in a physics-based simulator. We train and test our policy in various simulation environments. We benchmark our approach against other conventional occupancy data. The results show that our method not only decreases the required training time, but also increases the success rate of the navigation. As a future work, we will estimate our reward function as a neural network trained with an expert data in order to adapt our approach in more challenging real-world applications.








\bibliographystyle{IEEEtran}
\bibliography{main}
\end{document}